# Culture Matters in Toxic Language Detection in Persian

**Zahra Bokaei**
School of Informatics
University of Edinburgh
zahra.bokaei@ed.ac.uk

**Walid Magdy**
School of Informatics
University of Edinburgh
wmagdy@ed.ac.uk

**Bonnie Webber**
School of Informatics
University of Edinburgh
bonnie.webber@ed.ac.uk

## Abstract

Toxic language detection is crucial for creating safer online environments and limiting the spread of harmful content. While toxic language detection has been under-explored in Persian, the current work compares different methods for this task, including fine-tuning, data enrichment, zero-shot and few-shot learning, and cross-lingual transfer learning. What is especially compelling is the impact of cultural context on transfer learning for this task: We show that the language of a country with cultural similarities to Persian yields better results in transfer learning. Conversely, the improvement is lower when the language comes from a culturally distinct country.

**Warning:** This paper contains toxic language examples used solely for research on toxicity detection.

## 1 Introduction

Toxic language detection focuses on identifying and mitigating harmful content in text, including but not limited to hate speech, harassment, and threats (Hoang et al., 2024). With the rapid growth of online platforms and forums, the prevalence of such toxic language has become a pressing concern. Engaging in online discussions on social media, blogs, or comment sections often exposes users to hostile or disrespectful interactions (Olteanu et al., 2018). Such toxic behaviors not only undermine the overall quality and inclusivity of online communities but are also deeply intertwined with cultural and linguistic norms. What is considered toxic or inappropriate varies significantly across cultures (Sap et al., 2021; Zhou et al., 2023b), adding complexity to the task of automatic detection.

Over the years, studies have explored various techniques for tackling the challenge of detecting toxic language across diverse languages (Abro et al., 2020; Zimmerman et al., 2018; Badjatiya et al., 2017; Gaydhani et al., 2018). Since Large Language Models (LLMs) have demonstrated outstanding performance across diverse language-related tasks in multiple languages, there is increasing interest in assessing their effectiveness in detecting toxic content. (Khondaker et al., 2023; Kumar et al., 2024; Abaskohi et al., 2024).

However, toxic language detection in Persian remains under-explored, primarily due to the lack of high-quality datasets and tailored tools. Persian (also known as Farsi) and its variants—Dari and Tajik—are spoken by over 110 million people worldwide, with significant linguistic and cultural importance[1]. Addressing the challenges of toxic language detection in Persian is crucial due to its widespread use and the complexities introduced by its non-Latin script, diverse writing styles, and regional dialects. Only a recent work by Delbari et al. (2024) showed that advanced models, such as chat-GPT, struggle with detecting hate-speech in Persian, while the best performance using a fine-tuned Persian BERT model achieves only 0.61 F-Score.

To bridge this gap, our study investigates various approaches for Persian toxic language detection, including fine-tuning multiple LLMs, data enrichment, zero-shot and few-shot learning, and cross-lingual transfer learning. A key insight from our work is highlighting the critical role of cultural context in enhancing transfer learning effectiveness. Our findings indicate that models trained on languages of countries with greater cultural similarities to Persian achieve superior performance in detecting toxic content compared to those trained on large-scale English datasets, which offer only marginal improvements. This highlights the role of cultural similarities in improving model effectiveness, especially in context-dependent approaches.

The current study addresses four research questions (RQs):

RQ1. What is the performance of existing generative

---

[1] https://www.ethnologue.com/

LLMs on toxic language detection in Persian, using zero-shot and few-shot learning?
RQ2. Can fine-tuning enhance the performance?
RQ3. Would data enrichment via distant supervision improve Persian toxic language detection?
RQ4. Given that toxic speech classifiers are culturally insensitive (Lee et al., 2023), can cross-language transfer learning improve performance? Which languages perform best?

We explore these research questions through experiments on the PHATE dataset (Delbari et al., 2024), which includes three categories of toxic language in Persian: hate, vulgarity, and violence. For consistency, we utilize the same training, validation, and test splits as provided in PHATE. We find that toxic language identification in Persian continues to be a challenging task for most existing LLMs. However, between ParsBERT (Farahani et al., 2021) and Dorna2-Llama3 Instruct (PartAI, 2024), the two models specifically trained on Persian, Dorna2-Llama3 Instruct yield better overall results, also outperforming other multilingual models such as XLM-R and mT5. In addition, using distant supervision to obtain additional Persian training data significantly enhances the performance of ParsBERT compared to other models. We also find that transfer learning for Persian toxic detection is highly dependent on cultural context. Specifically, when the source and destination languages originate from culturally overlapping countries, the results tend to improve significantly.

## 2 Related Work

### 2.1 Toxic Language Detection

Early toxic language detection research focused on Machine Learning (ML) and Deep Learning (DL) techniques for English hate speech on social media (Asogwa et al., 2022; Davidson et al., 2017; Mullah and Zainon, 2021; Malik et al., 2024; Zimmerman et al., 2018; Zhou et al., 2020; Roy et al., 2020; Zhang et al., 2018), alongside efforts in offensive language (Bade et al., 2024; Aiyanyo et al., 2020; Cao et al., 2020; Risch et al., 2020) and cyberbullying detection (Wang et al., 2020; Pamungkas and Patti, 2019; Van Hee et al., 2015; Guo and Gauch, 2024; Cano Basave et al., 2013). Research has since expanded to languages like Indonesian (Ibrohim and Budi, 2019), Danish (Sigurbergsson and Derczynski, 2020), Arabic (Mubarak et al., 2021; Bensalem et al., 2023), Korean (Jeong et al., 2022), Chinese (Deng et al., 2022), Greek (Pitenis et al., 2020), and Indic languages (Gupta et al., 2022), including Hindi (Kapoor et al., 2019).

With LLMs, benchmarking across languages has further advanced the field (Zampieri et al., 2020; Verma et al., 2022; Caselli et al., 2021; Saleh et al., 2023; Nguyen et al., 2023; Chiu et al., 2021; Zampieri et al., 2023). Studies such as (Vargas et al., 2023), (Lu et al., 2024), and (Hoang et al., 2024) have demonstrated promising results for English. Shared tasks, like SemEval OffensEval (Zampieri et al., 2019), HASOC (Mandl et al., 2019), GermEval(Wiegand et al., 2018), OSACT5 (Mubarak et al., 2022), have fostered collaboration and innovation in this field.

However, research on Persian toxic language detection remains rare. Existing studies (Jey et al., 2022; Sheykhlan et al., 2023; Safayani et al., 2024; Ataei et al., 2023) provide limited publicly available datasets and primarily focus on a single category of toxic language. Recently, Delbari et al. (2024) provides a hierarchical, multi-label dataset categorizing violence, hate, and vulgarity, which forms the foundation of our work. The study evaluated different models, including ParsBERT, mBERT, XML-R, and ChatGPT, with the F1-Macro of 57.8, 55, 58.3, and 43.5 respectively. Because this work uses a limited dataset, relies solely on fine-tuning BERT-base models, with GPT models restricted to zero-shot scenarios, focuses only on binary classification tasks, and lacks thorough error analysis, we aim to address these limitations by enhancing the dataset with distant supervision, experimenting with various LLMs and transfer learning techniques considering the role of cultural similarities and expanding from binary to multi-class classification to better capture real-world complexities. Additionally, we establish a robust benchmark and perform comprehensive error analysis, offering deeper insights and a more reliable evaluation framework.

### 2.2 Transfer Learning

Transfer learning leverages pre-trained models to improve performance on new tasks with limited data. Understudied languages can benefit significantly from this technique, as pre-trained models provide a strong foundation for adaptation and learning (Unanue et al., 2023), even though they may yield suboptimal results for tasks that rely heavily on context and culture, (Zhou et al., 2023b). Bigoulaeva et al. (2021) uses cross-lingual transfer learning for hate speech detection, leveraging

English as the source and German as the target language. The approach successfully achieves strong performance on the target language without requiring annotated German data. Another study (Zhou et al., 2023a) focuses on detecting offensive language in Chinese using transfer learning with data from English and Korean. It finds that culture-specific biases hinder effective transferability.

### 2.3 Weak Supervision Annotation

Distant supervision is a weak supervision method that automates the creation of labeled training data by aligning unstructured text with existing annotated data. Magdy et al. (2015) demonstrates how distant supervision can assign YouTube video categories as labels to tweets linking those videos, enabling the generation of a large, automatically labeled dataset. Similarly, Go et al. (2009) applied this method for Twitter sentiment classification, achieving promising results. Additionally, studies such as (Lin et al., 2022), (Zeng et al., 2015), (Purver and Battersby, 2012), and (Mintz et al., 2009) have successfully deployed distant supervision across various NLP tasks, further showcasing its effectiveness. In this study, we introduce, for the first time in Persian, a novel distant supervision method to enhance the existing dataset.

## 3 Dataset

PHATE dataset, (Delbari et al., 2024), used in our study, consists of 7,056 tweets distributed across four classes: 582 labeled as violence, 1,583 as vulgar, and 1,632 as hate, with the remaining 3,259 categorized as neutral. The annotation methodology adopted in the baseline defines 'hate speech' as any instance labeled under vulgarity, violence, or hate, resulting in overlapping labels. Since our goal is distinct multi-class categorization rather than binary classification, we removed this overlapping label to concentrate on distinct toxic categories. We evaluate all models on the same test set and adhere to the baseline train-test-validation split (50-40-10) for comparability.

To apply distant supervision, we first needed to construct a Persian toxic lexicon. To this end, three native Persian speakers meticulously analyzed the training dataset to identify keywords frequently used in each category. The initial review identified 164 keywords, which were refined to 127 by removing terms—such as specific names—that could plausibly occur in neutral contexts, in order to reduce potential bias. The final selection was determined through majority voting among the annotators. At this point, nearly 40% of the keywords were associated with vulgarity.

We then followed a structured approach for each toxic class to further expand the lexicon. To enrich the "hate" category, we relied on definitions from the baseline annotation guidelines (Delbari et al., 2024) and introduced annotators to the most common hate targets. To do so, in addition to the hate targets identified by Silva et al. (2016), including racial and ethnic, religious, gender, individuals with disabilities, and other social groups, we added another hate target—politics—as the frequency of this target is reported to be high in the dataset (Delbari et al., 2024). Inspired by (Grimminger and Klinger, 2021), we also selected specific critical cultural events and asked annotators to generate keywords associated to hate speech based on those events. This approach ensured more contextually relevant hate speech categories, tailored to the sociocultural climate of the region. Annotators were asked to add relevant keywords associated with these targets, leaving categories blank where no suitable terms were identified. This process produced 216 distinct keywords, which were then narrowed down to 118 through majority voting. Next, for "violence" category, the annotators used the baseline definitions to identify relevant terms, ultimately finalizing 81 distinct keywords. Since the vulgarity class already had substantial representation, we supplemented it with 51 additional keywords at this stage.

To enhance the lexicon, we use the FastText model (Bojanowski et al., 2017) trained in Persian to identify related and synonymous terms for the 377 keywords identified earlier. Filtering out duplicates and irrelevant words, yielded a final lexicon of 604 toxic keywords across the three categories.

Using this toxic lexicon and a Twitter archive[2], containing tweets from 2011 to 2022, we identified tweets that included the identified toxic keywords. These tweets were then labeled according to the respective categories in our lexicon. To ensure our dataset remained distinct from the baseline dataset, which covers tweets from 2020 to 2023, we excluded any duplicate tweets from this overlapping period prior to the annotation process.

---

[2] https://archive.org/details/twitterarchive

| Model | #Params | Reference |
|---|---|---|
| ParsBERT | 162M | (Farahani et al., 2021) |
| XLM-RoBERTa-Base | 125M | (Conneau, 2019) |
| mT5-Base | 120M | (Xue et al., 2021) |
| Llama 3-Base | 8B | (Dubey et al., 2024) |
| Llama 3 Instruct | 8B | (Dubey et al., 2024) |
| Dorna2-Llama3.1-Instruct | 8B | (PartAI, 2024) |
| GEMMA 2 | 9B | (tea, 2024) |
| GPT 3.5 Turbo | 175B | (Brown, 2020) |

Table 1: LLMs used in our Study.

Ultimately, this process yielded 3291 toxic tweets across the three categories. To keep the dataset fairly balanced, we supplemented this with 3,200 neutral tweets. Tweets were considered neutral if they did not contain any of the toxic keywords from our lexicon.

## 4 Experimental Setup

Based on the results of the recent study by Delbari et al. (2024), we selected ParsBERT (Farahani et al., 2021) as our baseline model, as it has demonstrated promising results across a variety of Persian NLP tasks. Table 1 lists LLMs used in our benchmarking process. All models were trained for 10 epochs on PHATE, and the final results on the baseline test dataset are from the epoch with the highest F1 score on the validation set. This methodology ensures that we capture each model's optimal performance during evaluation.

### 4.1 Zero/Few shot Experiments:

In our experiments, we conducted few-shot and zero-shot evaluations with Llama 3 and GEMMA 2. However, due to their poor and non-competitive performance, we excluded these results from the benchmark. We employed GPT 3.5 Turbo in both zero-shot and few-shot settings to compare performance across each class, and a binary classification setting to evaluate whether the model performs better in binary or multi-label tasks. Inspired by prior work (Abaskohi et al., 2024), we exclusively used English prompts, as they have consistently demonstrated better performance for various Persian tasks. Our prompt provides definitions for each label, based on the definitions presented in (Delbari et al., 2024), which are partially derived from Twitter's rules and policies. The prompts used for our experiment is illustrated in Figure 5 in the Appendix.

### 4.2 Fine-tune Experiments:

We fine-tuned different LLMs on the enriched and baseline train datasets and evaluated their performance on the baseline test set, to maintain comparability. This allowed us (1) to assess the effectiveness of our distant supervision method in enriching the toxic dataset, and (2) to benchmark the performance of different state-of-the-art LLMs on the task of toxic content detection in Persian. Among our experiments on multilingual LLMs, Llama 3 consistently achieved better results than other models. Motivated by these findings and inspired by (Abaskohi et al., 2024), we conducted an additional experiment by translating the baseline training dataset into English using the Google Translate API. We then fine-tuned Llama 3 on the translated dataset and evaluated its performance using the baseline test set to analyze the impact of language translation on classification results. This step underscores Llama 3's adaptability and robustness across different languages.

### 4.3 Transfer Learning Experiments

Regarding transfer learning, we utilized three languages—Arabic, Indonesian, and English— and explored the interplay of linguistics and cultural factors in toxic speech detection. Since Llama 3 consistently achieved better results than other multilingual models, we selected this model for all our transfer learning experiments.

Arabic, a Semitic language, is commonly used for communication throughout the Arab world. It is written in the Arabic script and is known for its rich structure, complex grammar, and variety of regional dialects. Arabic was included in this study due to its cultural and linguistic similarities with Persian, as both languages share certain linguistic and cultural features and use similar scripts.

English, a high-resource language with extensive datasets, allows us to assess how effectively models can adapt knowledge from a linguistically and culturally unrelated yet well-documented source.

Indonesian, or Bahasa Indonesia, is the official language of Indonesia and a standardized form of Malay. As part of the Austronesian language family, it is spoken by millions across the Indonesian archipelago. Indonesian was selected for this study due to its cultural ties with Persian, enabling an exploration of how cultural similarities and linguistic differences impact transfer learning.

Regarding Arabic, we leverage the availability of large datasets for vulgar and hate speech (Mubarak et al., 2022) to examine whether the cultural and linguistic proximity between Arabic and Persian supports this approach. In one experiment, we train the Llama 3-base model on Arabic vulgar and hate datasets and evaluate its performance on the baseline test set. In another experiment, we combine the baseline Persian training dataset with the Arabic dataset, retrain the Llama 3-base model, and test it on the baseline test set. A similar approach has been applied to English, leveraging extensive datasets containing hate, vulgarity, and violence (Kennedy et al., 2020), as well as to Indonesian, utilizing a comprehensive hate dataset (Ibrohim and Budi, 2019). Finally, we conducted two additional experiments: first, by combining the Indonesian and Arabic datasets to retrain the Llama 3 base model and evaluating it on the baseline test set; and second, by integrating the baseline training dataset with the Indonesian and Arabic datasets and repeating the experiment. To ensure comparability, we used datasets of equal size for all languages while maintaining a balanced label distribution across all classes. We achieved this by randomly selecting an equal amount of data from each dataset. Due to the absence of large datasets for vulgar or violent language, our Indonesian experiments focused solely on hate detection. Similarly, the lack of Arabic dataset for violence restricted our transfer experiments to English. Table 2 shows the distribution of the data set and the selection of subsets for hate classification in transfer learning.

| Language | Dataset size | | | Random subset used | | |
|---|---|---|---|---|---|---|
| | Total | Hate | Not Hate | Total | Hate | Not Hate |
| English | 39,565 | 10,892 | 28,673 | 8,050 | 4,025 | 4,025 |
| Indonesian | 13,169 | 5,561 | 7,608 | 8,050 | 4,025 | 4,025 |
| Arabic | 12,698 | 4,025 | 8,673 | 8,050 | 4,025 | 4,025 |

Table 2: Dataset distribution and subset selection for hate classification in transfer learning.

## 5 Results

This section is divided based on the results obtained using different methods as Zero-Shot/Few-Shot, Fine-Tuning, Distance Supervision, and Transfer Learning approach. Table 3 presents a comprehensive comparison of model performance.

### 5.1 GPT 3.5 Turbo Few-Shot and Zero-Shot

For multi-class classification, GPT 3.5 Turbo - 0 Shot achieved moderate scores across categories, while GPT 3.5 Turbo - 2 Shot improved these met-rics, notably for Hate and Violence. However, increasing the number of shots beyond two did not yield significant improvements in performance. To optimize resource utilization, we limited our experiments to 2-shot settings for multi-class classification and shifted our focus to binary classification for further evaluation. In binary classification, models demonstrated significantly higher performance overall. GPT 3.5 Turbo - 0 Shot achieved top scores in categories such as "Violence" and "Hate".

| | Model | Violence | | | Vulgar | | | Hate | | | $F_m$ |
|---|---|---|---|---|---|---|---|---|---|---|---|
| | | P | R | $F_1$ | P | R | $F_1$ | P | R | $F_1$ | |
| Zero/Few shot | GPT 0-shot | 35 | 75 | 48 | 61 | **46** | 52 | 39 | **89** | 54 | 51 |
| | GPT 2-shot | 40 | **81** | 54 | 79 | 37 | 50 | 55 | 69 | 61 | 55 |
| | GPT 0-shot binary | <u>**81**</u> | 73 | <u>**77**</u> | **85** | 30 | 44 | **83** | 64 | 72 | 64 |
| | GPT 1-shot binary | 80 | 70 | 75 | 74 | 43 | 54 | 77 | 83 | **80** | 69 |
| | GPT 2-shot binary | 78 | 75 | 76 | 77 | 42 | **55** | 74 | 86 | **80** | **70** |
| | GPT 3-shot binary | 79 | 71 | 75 | 76 | 36 | 49 | 76 | 81 | 78 | 67 |
| Fine tuning | ParsBert (Baseline) | 68 | 42 | 52 | 55 | **68** | 60 | **63** | 59 | 60 | 57 |
| | Dorna2-Llama Inst. | 61 | **74** | **67** | 50 | 52 | 55 | 56 | 73 | 60 | **61** |
| | XLM-R-base | 63 | 50 | 56 | 55 | 63 | 59 | 58 | 67 | **62** | 59 |
| | Llama 3 - Base | 68 | 57 | 62 | 51 | 65 | 57 | 53 | **76** | **62** | 60 |
| | Llama 3 translated | 48 | 57 | 52 | 36 | 34 | 35 | 49 | 67 | 57 | 48 |
| | Llama 3 Instruct | **74** | 55 | 63 | 58 | 57 | 57 | 59 | 55 | 57 | 59 |
| | GEMMA 2 | 57 | 35 | 43 | 40 | 54 | 46 | 51 | 69 | 59 | 49 |
| | mT-5 | 38 | 41 | 39 | **59** | 26 | 36 | 56 | 49 | 52 | 42 |
| Distant supervision | ParsBert | **62** | 58 | 60 | **78** | 67 | **72** | 71 | **81** | **75** | **69** |
| | XLM-R | 54 | 69 | **61** | 76 | 63 | 69 | **71** | 74 | 72 | 67 |
| | Llama 3 | 36 | **70** | 47 | 56 | 51 | 53 | 70 | 57 | 63 | 54 |
| | Gemma 2 | 37 | 65 | 47 | 44 | 50 | 47 | 64 | 54 | 58 | 51 |
| | mT-5 | 34 | 61 | 44 | 52 | 62 | 57 | 45 | 74 | 56 | 52 |
| Transfer learning | Llama 3 - En | 78 | 69 | 73 | 74 | 81 | 77 | 55 | 60 | 57 | 69 |
| | Llama 3 - En+Fa | 79 | 70 | 74 | 81 | 78 | 80 | 56 | 61 | 59 | 71 |
| | Llama 3 – Ar | - | - | - | 81 | <u>**84**</u> | 82 | 75 | 89 | 81 | 82 |
| | Llama 3 – Ar+Fa | - | - | - | **83** | <u>**84**</u> | **84** | 86 | 88 | 87 | 86 |
| | Llama 3 – Id | - | - | - | - | - | - | 89 | 84 | 86 | 86 |
| | Llama 3 – Id+Fa | - | - | - | - | - | - | 92 | 80 | 86 | 86 |
| | Llama 3 – Ar+In | - | - | - | - | - | - | <u>**94**</u> | <u>**92**</u> | <u>**93**</u> | <u>**93**</u> |
| | Llama 3 – Ar+In+Fa | - | - | - | - | - | - | 92 | 91 | 91 | 91 |

Table 3: Toxic detection across all approaches. Best in each group is in bold. Overall best is underlined. $F_m$ is macro F-score across the existing toxic categories.

### 5.2 Fine Tuning

The fine-tuning results revealed distinct trends among the four LLMs groups.

**BERT Models:** ParsBERT, the BERT-base model, served as the baseline (Delbari et al., 2024) achieved moderate F1 scores for all categories. When fine-tuned with an enriched dataset, ParsBERT with Distant Supervision showed significant improvements on the baseline test set, particularly for "Hate" (F1 = 75) and "Vulgar" (F1 = 72). Additionally, the performance of the XLM-R-base model, fine-tuned with the enriched dataset, improved significantly across all categories.

**Llama Models:** The Llama models displayed varied performance depending on the dataset and specefic models. Llama 3 – Base, trained on the baseline dataset, achieved F1 scores of 62, 62, and 57 for "Violence," "Hate," and "Vulgar," respectively. However, its enriched counterpart, Llama

3 with Distant Supervision, showed mixed results: while the F1 score for "Hate" improved, the score for "Violence" dropped significantly, highlighting challenges in effectively utilizing enriched datasets. A similar drop occurred for "Vulgar." Compared to other models, Llama 3 – Translated, fine-tuned on English-translated baseline dataset, underperformed, suggesting that translation into English may have removed critical linguistic features necessary for effective classification. Llama 3 – Instruct trained on the baseline training dataset achieved consistent F1 scores of 63, 57, and 57 across the three categories. Building on these findings, we extended our experiments by incorporating the recently released Dorna2-Llama3 Instruct, which outperformed both Llama 3 – Instruct and Llama 3 – Base, achieving higher F1 scores for the 'Violence' and 'Hate' classes. Notably, among all fine-tuned models in our experiments, this model achieved the highest results for detecting 'Violence'.

**GEMMA Models:** The GEMMA 2 models, underperformed compared to Bert - base and Llama - base models. Enriching the dataset offered marginal improvements for "Vulgar" but for "Violence" increased 4% and "Hate" dropped by 1%. These results highlight the limitations of GEMMA 2 in task-specific Persian contexts.

**mT-5 Model:** mT-5 exhibited the weakest performance among all fine-tuned models. While mT-5 with Distant Supervision showed slight improvements, it struggled to achieve competitive results.

### 5.3 Transfer Learning

We observed that fine-tuning on English data alone (Llama 3 – Eng) yielded moderate results: While the model performed well in "Violence" and "Vulgar," its performance in "Hate" was weaker. Including Persian in the training process alongside English (Llama 3 – Eng + Fa) improved the F1 scores across all categories.

Fine-tuning on Arabic data alone (Llama 3 – Ar) yielded strong F1 scores of 81 for both "Hate" and "Vulgar." Adding Persian data (Llama 3 – Ar + Fa) further enhanced performance, with F1 scores of 87 for "Hate" and 84 for "Vulgar."

Fine-tuning on Indonesian alone (Llama 3 – Id) resulted in an F1 score of 86 for 'Hate.' However, incorporating Persian data into the Indonesian training set (Llama 3 – Id + Fa) further improved precision while maintaining a consistent F1 score.

Integrating both Arabic and Indonesian datasets (Llama 3 - Id + Ar) achieved the highest F1 score of 93 across all experiments. However, adding Persian (Llama 3 - Id + Ar + Fa) resulted in a slight decrease, bringing the F1 score down to 91.

## 6 Analysis and Discussion

### 6.1 RQ1: Generative LLMs Performance

Our first RQ concerned the performance of existing generative LLMs, using zero-shot and few-shor learning: We observed that in zero- and few-shot settings, GPT-3.5 Turbo performs significantly better in binary classification tasks than in multi-label classification. In zero-shot multi-label classification, the model frequently mislabeled instances, often confusing categories such as 'hate' and 'violence.' Additionally, some instances of 'hate' are incorrectly classified as 'neutral,' particularly when lacking sufficient contextual cues.

Analysis of few shot multi-label classification reveals misclassifications that even though they contain elements of vulgarity or violence such as keywords like "down with" and "dead to", or discussions about public figures and specific locations do not meet the criteria for hate speech. Moreover, as in zero-shot multi-label classification, some instances of 'hate' are misclassified as 'neutral,' especially those related to specific events. Table 4 shows some GPT 3.5 Turbo misclassified samples.

Given GPT 3.5 Turbo's stronger performance in binary settings, we conducted three few-shot experiments with 1-shot, 2-shot, and 3-shot settings, with noticeably better performance. After analyzing the errors in the binary setting, we found that GPT-3.5 Turbo similar to multi-classification experiments relies heavily on contextual clues in the text to distinguish between these labels. However, the predictions can skew incorrectly when the context is ambiguous or conceptually overlapping. For example, while the model successfully detects hate with common targets (e.g., religion, politics), it struggles to detect hate for targets related to specific events. Table 5 presents some of these misclassifications. Interestingly, the model's performance either remained steady or dropped as the number of shots increased. Analysis reveals that instances relying on context struggle to predict correctly, even in a 3-shot setting. This finding aligns with prior work that conducted exhaustive experiments on GPT models across various Persian tasks (Abaskohi et al., 2024). For N-shot binary and multi-class classification, we tested various instances at each level and selected the average-performing outputs for reliability.

| Tweet (original + English translation) | Actual Label | Predicted Label | | | | |
|---|---|---|---|---|---|---|
| | | 0-shot multi | 0-shot binary | 1-shot binary | 2-shot binary | 3-shot binary |
| گفتگو؟؟ سه ساله هرروز دارم میپرسم #موشک_دوم رو چرا زدید<br>Conversation??? For three years, we've been asking every day why you fired the second missile. | Hate | Violence | Neutral | Neutral | Neutral | Neutral |
| دختری جوان برای عمل جراحی زیبایی به کلینیکی مراجعه میکند و زیر تیغ سکته میکند؛ جسد او را به خارج برند و آن را آتش زدند. نمیخواهید کل هیکل سازمان نظام پزشکی را از بالا تا پایین اقاله بگیرید؟<br>A young girl visits a clinic for cosmetic surgery and suffers a stroke under the knife; her body is taken abroad and set on fire. Don't you want to take the whole Medical System Organization from top to bottom and throw it in the trash? | Hate | Vulgar | Hate | Hate | Neutral | Neutral |
| خدا رو شاکرم که علیرغم پذیرش در آزمون قضاوت و گزینشهای مربوطه به شغل شریف قضاوت نائل نیامدم تا مجبور نباشم زمانی که پدر دو کودک ۸۰ روز در بازداشت انفرادی به سر میبرد حکم مادر آنها نیز بدهم!<br>I thank God that despite being accepted in the judicial exam and the related selections, I did not attain the honourable position of a judge, so I wouldn't have to give a verdict to detain the mother of two children while their father spends 80 days in solitary confinement! | Hate | Neutral | Neutral | Neutral | Neutral | Neutral |

Table 4: Samples of Hate Misclassifications by the GPT Binary/Multi Classification Experiment.

### 6.2 RQ2: Fine-Tuning Effect

Our Second RQ concerned fine tuning: Regarding models specifically trained on Persian, in comparison to others, ParsBERT still lagged in detecting toxic language. In contrast, the recent Dorna2-Llama3.1-Instruct achieved better overall results.

Regarding multilingual LLMs, Llama 3 performs better than GEMMA 2, with mT5 being the worst among them. We also used Llama-Instruct with a definition of the classification task, but did not observe significant differences in performance. Using the translated dataset, we observed that all metrics dropped notably: likely due to the problematic translations, As most entries were informal and context-dependent, they were difficult for Google Translate to process correctly.

### 6.3 RQ3: Data Enrchiment via Distant Supervision

Our third RQ concerned the effect of data enrichment via distant supervision: Our results demonstrate that distant supervision improves mT5 and significantly enhances BERT base models. However, it performs poorly on Llama 3 and GEMMA 2. The metrics reveal that the results on Llama 3 are 50% worse than those on GEMMA 2, suggesting that Llama 3 is less tolerant to noise when trained on Persian. Additionally, our proposed dataset introduces a drop in precision for detecting violence across all models.

As highlighted by (Magdy et al., 2015), distant supervision, despite its inherent noise, can substantially enhance model performance by providing additional contextual data during training. This observation aligns with our findings, where the BERT-base models demonstrated improved performance with distant supervision.

However, as Table 3 shows, for ParsBERT and XLM-R, the precision for the "violence" category dropped by an average of 7%. A detailed analysis of misclassified labels revealed that 68% of "neutral" labels were erroneously classified as "violence." This misclassification primarily stemmed from overlapping keywords and contextual ambiguities triggered by our toxic lexicon. For example, in the enriched dataset, the word بمیر (kill) often appears in both "neutral" and "violent" contexts. While in Persian it is typically used humorously or exaggeratedly in neutral conversations, the models frequently misclassified it as "violent". Similarly, terms like موشک زدند (barrage rocket) and منفجر

(explode), neutral in certain contexts, were incorrectly labeled as violence. Table 6 displays some of the false positive instances resulting from the model. Since most of these tweets were correctly labeled as neutral during the baseline training of the BERT-base models, this suggests that our distant supervision method introduced noise, complicating the differentiation between categories in this context.

In addition, we observed that, although the instances for the "vulgar" category increased by approximately 40% through distant supervision, the recall remained almost unchanged for both ParsBERT and XLM-R. This stability in recall suggests that the additional data introduced by distant supervision might not have been sufficiently diverse or contextually rich to enhance the models' performance. Moreover, the models still struggle with implicit profane speech. Table 6 presents instances that were not detected as 'vulgar' during training on both datasets, even though they explicitly contain words from our toxic lexicon.

In contrast, our dataset significantly improves the recall for "hate". We observed that this is especially true for hate directed towards politics, where the model trained on the baseline dataset struggled to identify instances. However, after training on the enriched dataset, it successfully detected these instances, suggesting that our approach for identifying hate keywords in the toxic lexicon works well for hate detection.

| Tweet (original + English translation) |
|---|
| رشتو در کشورهایی نظیر میهن ما ایران که مردم یا بحقوق مدنی خودشان واقف نیستند یا ایستادگی ندارند بگزریم که از آن پای ایستادگی بگزریم که از آن گروه بیشماری که با ساندیس و ساندویچی به با آن حقوق چشم میپوشند و فریادهای مرگ بر ... یا الله اکبر خامنه ای رهخر سرمیدهند، صحبت جمهوری خواهی پس از براندازی k<br>The thread in countries like ours, Iran, where people are either unaware of their civil rights or do not stand firm to claim them — let alone the countless groups who, in exchange for a juice and a sandwich, turn a blind eye to those rights and chant slogans like **'Death to...'** or 'Allah Akbar Khamenei Rahbar' — speaks of republicanism after the overthrow k. |
| به نشمان #عرفانی_پور و #حسن_عباسی میگم مناظره میاد میگن احمق نیستیم که وقتمون را با حرفای اینا اهمیت نمیبینم!😂 بعد از این طرف نوبه نوبه سخنرانی هاشون آنلاین میکنن و از ز سخنرانیای صندسال پیششون فلان مثلا سوتی رو درمیارن یا فلان حرف و انه شکایت ...<br>##**Rafi_poor** and **#Hassan_Abbasi** are in the midst of an idiotic debate when they come to me with a black face that literally means n o t h i n g ! ! 😂 after that party Don't say anything about someone, for example, say something about someone, for example, say something like a letter with a c o m p l a i n t 😂 |
| نیمار اومده به بازیکن ژاپنی مارسی توهین نژادی بکنه به به جای اینکه بگه بگه ژاپنی گوه گفته چینی گوه قشنگ ریده تو کل آسیا. به غیر از اون به بازیکن مارسی هم همچنگرا خطاب کرده که احتمالا محرومیت سنگینی رو در انتظار داره. متاسفانه پول علاوه بر سواد جغرافیایی هم نمی تون داشت<br>Neymar insulted the **Japanese Marseille** player with a racial slur. Instead of saying 'Japanese shit,' he said **'Chinese shit,'** completely screwing over all of Asia. Besides that, he also called the Marseille player gay, which will likely result in a heavy suspension. Unfortunately, money can't buy intelligence or geographical knowledge. |
| رژیم صهیونیستی در واقع یک رژیمی است که پایه های سست است، رژیم صهیونیستی محکوم به زوال است #freepalestine #مهد_مقاومت<br>**The Zionist regime** is a regime with extremely fragile foundations; the Zionist regime is doomed to collapse.#freepalestine #Stronghold_of_Resistance |

Table 5: Samples of false negative classifications by the GPT for the Hate class.

| Tweet (original + English translation) | Actual Label | Predicted Label |
|---|---|---|
| زیر بارون باهم قدم بزنیم تو چترتو واسه من نگه داری که من خیس نشم ولی خودت زیر بارون خیس بشی بعد تو سرما بخوری کرونا بگیری بمیری که وقتی میگم بیا بریم  خونه نگی نه بریم قدم بزنیم (((:<br>Let's walk together in the rain, and you hold the umbrella over me so I don't get wet, but you get soaked in the rain. Then you catch a cold, get COVID, and die, just so the next time I say, "Let's go home," you don't say, "No, let's keep walking." :)))) | Neutral | Violence |
| ای بمیری چقدر شکر بهش زدی<br>Ugh, die already! How much sugar did you add to it | Neutral | Violence |
| وقتی کابلهای برق نطنز اتصالی کنه خو مشخصه که مرکز موشک سازی اسرائیل منفجر میشه😂.<br>When the power cables in Natanz short-circuit, of course, the missile manufacturing centre in Israel is going to explode. | Neutral | Violence |
| عمل زیبایی نه مایه شرمه نه افتخار. (از مجموعه گه یکدیگر را نخوریم)<br>Cosmetic surgery is neither a source of shame nor pride. (From the "Let's Not Eat Each Other" collection) | Vulgar | Hate |
| همون سالی که یارو حادثه رو با سریال واکینگ دد و زامبی ها مقایسه کرد باید به عقش شک میکردید 2<br>The year that guy compared the incident to *Walking Dead* and zombies was the moment you should've questioned his sanity. | Vulgar | Neutral |

Table 6: Samples of misclassification after training ParsBERT on enriched dataset.

## 6.4 RQ4: Cross-Lingual Transfer Learning

Our fourth RQ concerned the effect of culture in transfer learning: Our findings indicate that while Persian can effectively benefit from the Arabic and Indonesian datasets, its performance gains from the English dataset are less pronounced. Closer analysis of the results suggests two potential reasons for this disparity. First, the general culture of hate in Persian, Arabic, and Indonesian appears to be more similar, particularly in targets related to religion, politics, and common controversial events that provokes hate. In contrast, the English hate dataset predominantly focuses on contexts diverging significantly from the Persian hate dataset (e.g. sexual orientation and ethnic groups). Second, both Persian and Arabic are morphologically rich languages. This shared characteristic can allow Persian to exploit the morphological richness of Arabic during transfer learning, leveraging the capacity of LLMs to process such linguistic features effectively. The pattern observed with the hate class was mirrored in the vulgar class, where Persian again benefited more from Arabic than from English. However, to assess whether the effectiveness is more cultural or linguistic, we experimented on Indonesian, which has completely distinct linguistic features from Persian. As the results show, despite its linguistic divergence, training solely on the Indonesian dataset produced even better results than Arabic. Interestingly, our experiments demonstrated that English can still provide relevant contextual information about violence applicable to Persian.

Integrating datasets from three language pairs (Arabic-Persian, English-Persian, and Indonesian-Persian) showed improved performance metrics in the first two settings, except for a slight decline in recall for the "vulgar" class in the English-Persian combination (3%) and the "hate" class in the Arabic-Persian combination (1%). These minor drops can likely be attributed to the imbalance in data samples between the two datasets (e.g. PHATE and Indonesian). Upon further examination, we observed that, the transfer learning experiments reveal some differences in how Arabic and Indonesian datasets contribute to Persian toxic language detection. Specifically, transfer learning from Arabic data helped detect hate speech related to religious and political topics, particularly sociopolitical hate prevalent in the Middle East. This indicates that Arabic dataset provides relevant contextual cues for religious and politic discourse. On the other hand, transfer learning from Indonesian data helped detect hate speech directed toward individuals rather than groups (e.g. profession). In addition, our analysis highlights that models trained on Indonesian data exhibit significantly better performance in handling long texts containing a mix of neutral and hateful sentiments. This can be one reason Indonesian outperforms Arabic in detecting Persian hate instances. Close analysis of the

| Tweet (original + English translation) | Ar | Ar+Fa | In | In+Fa | Ar+In | All |
|---|---|---|---|---|---|---|
| اسرائیلی ها اینقدر زیاده خواه و بی منطق اند که محمود عباس تهدید کرده است "چنانچه اوضاع تغییر نکند علیه رژیم صهیونیستی اقدام خواهیم کرد" <br> Israelis are so greedy and irrational that Abbas has threatened, "If the situation doesn't change, we'll take action against the Zionist regime." | 1 | 1 | 0 | 1 | 1 | 1 |
| رندی به محضر فقیهی رسیدن رقص رقص را جداجدا انجام می داد و می پرسید آیا حرام است؟ فقیه میگفت نه. پس رند شروع به رقصیدن کرد.فقیه گفت تجزیه اش خوب بود ولی مرده شور ترکیبش رو ببرن **حالا حکایت این عدالت خواراست بعضیشون عییبی بچه های خوبین ولی مرده شور ترکیبشون رو برده من برم به کار و کاسبی خودم برسم خداحافظ #عدالتخواران#انتخابات_مجلس** <br> A trickster went to a cleric and performed dance moves separately, asking if they were forbidden. The cleric said no. Then the trickster started dancing, and the cleric said, 'Breaking it down was fine, but damn the combination!' **This is exactly the case with these so-called justice-seekers—some of them are actually good kids, but damn their combination! Anyway, I'll get back to my own business. Goodbye. #JusticeSeekers #ParliamentElection** | 0 | 1 | 1 | 0 | 1 | 1 |
| علیرضا دبیر: صحبت راجب سیاست شرعاً مشکل داره از بروبچ کشتی میخوام گوشیشون رو کنار بزارن و رو تمریناتشون تمرکز کنن. بعد از انقلاب اولین نفری که با مهسا بهش تجاوز که تویی بی وجود# مهسا_امینی <br> Alireza Dabir: "Talking about politics is religiously problematic. I ask the wrestling guys to put their phones aside and focus on their training." After the revolution, the first person I'll get my dog violate you, you worthless being. #Mahsa_Amini | 0 | 1 | 1 | 1 | 1 | 0 |
| ازماست که برماست تا به این دین و باورهای بیابان گرد ملخ خوار باور داریم **همین آش و همین کاسه** <br> It is up to us to believe in this religion and beliefs of the desert, the locust-eating locusts, **the same soup and the same bowl.** | 0 | 1 | 0 | 1 | 0 | 1 |
| از دی به سینا ازسینا به نجمیه به نجمیه چه غلطی دارن میکنن ازنجمیه هم میخوان برن امیرعلم حتما بیشرفا <br> From Dey to Sina, from Sina to Najmieh, what the hell are they doing? Now they want to move from Najmieh Amir Alam. They must be absolute scoundrels | 0 | 0 | 0 | 0 | 0 | 0 |

Table 7: Transfer Learning model predictions for Hate Farsi tweets across multiple languages.

Indonesian dataset, showed that it lacks sufficient political hate speech instances, which explains the model's struggle to generalize to such cases in the Persian context.

Furthermore, both transfer learning approaches reveal challenges in detecting instances containing idiomatic expressions and culturally dependent references that require specific background knowledge. We observed that integrating Persian data into the training process helps mitigate these challenges for Arabic and Indonesian datasets, with a more pronounced improvement in the Arabic model. We aimed to explore whether incorporating Indonesian and Arabic datasets could improve Persian hate speech detection and whether these two languages complement each other in identifying Persian hate speech. Upon examination, we confirmed our hypothesis: the integration of these two languages effectively complemented each other, improving detection capabilities. However, when we integrated all available training data—Indonesian, Arabic, and Persian—and trained a model using this combined dataset, we observed a slight drop across all metrics, although the results remained strong. A closer error analysis failed to reveal a clear pattern explaining this decline. Further investigation is needed to determine why incorporating Persian did not lead to additional improvements. Table 7 provides sample predictions that support our findings. More examples present in Table 8 in the Appendix.

## 7 Conclusion

This paper presented a comprehensive evaluation of various fine-tuning, zero-shot/few-shot, and transfer learning methodologies to assess the performance of LLMs in detecting toxic content in Persian—a low-resource language. Given the limited availability of data for Persian, we explored distant supervision to enrich existing Persian datasets and transfer learning to evaluate Persian's ability to leverage resources from other languages.

Our analyses demonstrate that distant supervision significantly enhances the performance of BERT-based models, particularly ParsBERT. We also show that transfer learning appears more effective when the language belongs to a country with cultural similarities to Persian, whereas improvements appear less significant for languages from culturally distinct countries.

## Limitations

One limitation of our study is that the toxic lexicon introduced for distant supervision cannot comprehensively capture all forms of toxic speech. Additionally, some keywords in the lexicon are heavily event-specific and may lose relevance over time as those events fade from public memory. This limitation suggests that the lexicon may not effectively identify toxic language associated with future events that provoke toxicity. Furthermore, other forms of toxic speech, excluded due to dataset constraints, present opportunities for future research to improve toxic speech detection frameworks. While our study focuses on three languages, limiting broader conclusions about cross-lingual transfer learning, our selection was guided by cultural relevance to Persian. Arabic and Indonesian were chosen for their linguistic and cultural ties, while English served as a high-resource control language. Further studies should explore additional languages to enhance cross-lingual generalizability.

## Ethics Statement

This study follows ethical principles, emphasizing fairness, privacy, and harm prevention in toxic content detection. Methods aim to minimize bias and ensure transparency and accountability. All data were handled in line with ethical guidelines and data protection laws.

# A Appendix

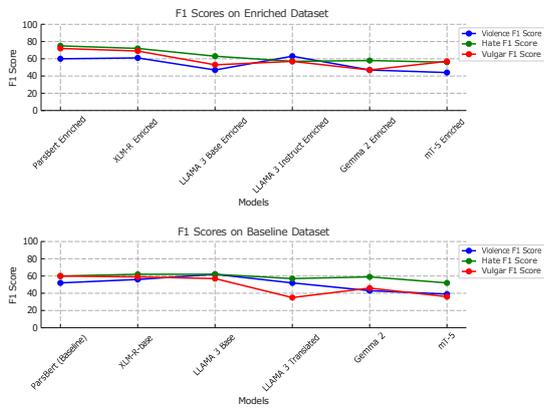

Figure 1: The fine-tuned models' performance before and after dataset enrichment.

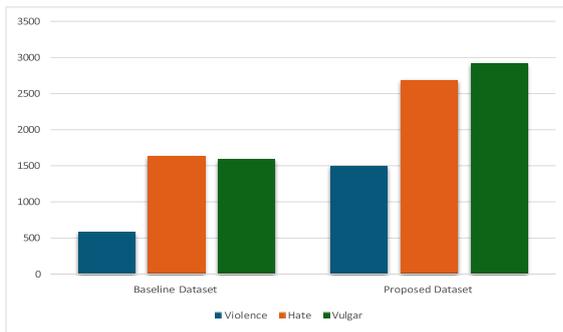

Figure 2: Label Distribution Before and After the Enrichment.

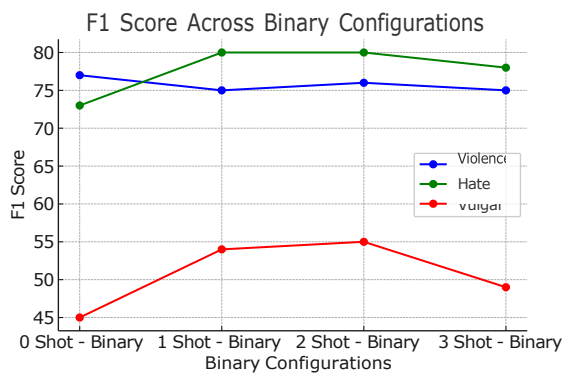

Figure 3: GPT F1 Score Results for Three Classes Across N-Shot Configurations.

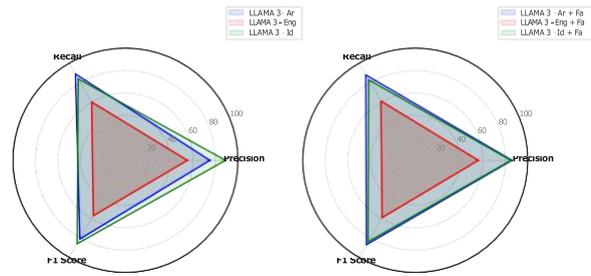

Figure 4: Performance of Transfer-Learning methods on the Test Set for Hate Speech Detection.

**Task Description:**
Classify the following text into one of two categories: Violent or Neutral.
Determine if the text meets the criteria for violence and assign the appropriate category.

**Label Description:**
1 (Violent):
The text is classified as Violent if it includes:
  1. Threats of violent acts against an identifiable target.
  2. Wishing, hoping, promoting, inciting, or expressing a desire for death or serious physical harm.
  3. Calling for and encouraging others to harm or harass.

0 (Neutral):
The text is classified as Neutral if it does not meet any of the criteria for being Violent.

**Example Pattern:**
<text>
<category>

**Examples:**
<text>: "با وجودی که امروز حسابی از نتیجه انتخابات ترسیدم ولی همه چیز خوب تموم شد".
<category>: 0

<text>: "آقای میرزایی یک بار دیگه به بازیکنان ایتالیا توهن کنی میام صدا سیما با همون مهدی توتونچی تا میخوری میزنم".
<category>: 1

Figure 5: The Prompt Used for the GPT Experiment.

| Tweet (original + English translation) | Ar | Ar + Fa | Ind | Ind + Fa | Type |
|---|---|---|---|---|---|
| بمب که بسازیم نه احتیاج به انتخابات دار نه هیچ فشاری از طرف داخل و خارج موشک هم دارم،سپاه هم دارم.گروههای نیابتی هم دارم،اسرائیل هم که آنجاست باج میگیرم و حکومت میکنم،مردم  هم غلط کرده اندکه به مُجتبی روی خوش نشان ندهند..جمهوری کره شمالی اسلامی<br>If we build a bomb, we won't need elections, nor will we face any internal or external pressure. I have missiles, I have the IRGC, I have proxy groups, and Israel is right there—I can extort and rule. And the people have no right to oppose Mojtaba. An Islamic North Korea | 0 | 0 | 0 | 0 | Implicit hate - Sarcastic |
| آغاز #دهه_زجر، آغاز اعدام بلند پایهترین مقامات کشوری و لشکری میهن پرست، آغاز اعدام مردم بیگناه و آزادیخواه، آغاز سالها فقر بدبختی گرانی تورم نداشتن فحشا نادانم کاری برای ایرانیان و آغاز  پایان امنیت #خاورمیانه را به تمامی دغدغهها و #جهاد_فکتوری تبریک و به #مردم_ایران تسلیت میگویم.<br>The beginning of the #Decade_of_Agony marks the start of the execution of the highest-ranking patriotic civil and military officials, the execution of innocent and freedom-loving people, the beginning of years of poverty, misery, inflation, prostitution, and incompetence for Iranians, and the start of the end of security in the #MiddleEast. Congratulations to the indifferent ones and #Jihad_Factory, and my condolences to the #People_of_Iran | 1 | 1 | 0 | 1 | Politics |
| اسرائیلی ها اینقدر زیاده خواه و بی منطق اند که محمود عباس تهدید کرده «چنانچه اوضاع تغییر نکند علیه رژیم صهیونیستی اقدام خواهیم کرد»<br>The Israelis are so greedy and irrational that Mahmoud Abbas has threatened, "If the situation does not change, we will take action against the Zionist regime." | 1 | 1 | 0 | 1 | Politics |
| تف تو مملکتی که دیه الناز رکابی از مهدی ترابی کمتره 🟡⚫<br>Shame on a country where Elnaz Rekabi's blood money is worth less than Mehdi Torabi's...! 🟡⚫ | 1 | 1 | 0 | 1 | Politics |
| رندی به محضر فقیهی رسیدو حرکات رقص را جداجدا انجام می داد و می پرسید آیا حرام است؟ فقیه میگفت نه. پس رند شروع به رقصیدن کرد،فقیه تجزیه اش خوب بود ولی مرده شور  ترکیبش رو بردن حالا حکایت این عدالت خواهان بعضیشان عینی بچه های خوبن ولی مرده شور ترکیبشون رو برده من برم به کارای خودم برسم خداحافظ #عدالتخواران#انتخابات_مجلس<br>A trickster went to a cleric and performed dance moves separately, asking if they were forbidden. The cleric said no. Then the trickster started dancing, and the cleric said, 'Breaking it down was fine, but damn the combination!' This is exactly the case with these so-called justice-seekers—some of them are actually good kids, but damn their combination! Anyway, I'll get back to my own business. Goodbye. #JusticeSeekers #ParliamentElections | 0 | 0 | 1 | 1 | Mix of Neutral and Hate |
| رفتم دماغمو عمل کنم دکتره یه نگاهی کرد گفت داداش شما صبر کن یکم دیگه علم پیشرفت کنه کلا سرتو  عوض کن🤦‍♂️؟ حالا شده حکایت ما مملکتی که رقابت  خوشگلترین مسئولش بین جهانگیری  و احمد خاتمی و احمدی نژاد باید رید توش البته از نظر  عملکرد شان.  هم باید رید توش #مهسا_امینی #ترور_بیولوژیکی<br>I went to get my nose done, and the doctor took one look at me and said, 'Bro, just wait a little longer until science advances enough to replace your whole head. 🤦‍♂️' That's exactly our situation—when the competition for the most handsome official in the country is between Jahangiri, Ahmad Khatami, and Ahmadinejad, you know it's doomed. And in terms of their performance and dignity, well, it's even worse. #Mahsa_Amini #Biological_Assassination | 0 | 1 | 1 | 1 | Mix of Neutral and Hate (bold) |
| علیرضا دبیر: صحبت راجب سیاست شرعا مشکل داره، از بروجی کشتی میخوام گوشیشون رو کنار بزارن و رو تمریناتشون تمرکز کنن. بعد از انقلاب اولین نفری که به میدم سگم تجاوز که  تویی بی وجود# مهسا امینی<br>Alireza Dabir: "Talking about politics is religiously problematic. I ask the wrestling guys to put their phones aside and focus on their training." After the revolution, the first person I'll have my dog violate is you, you worthless being. #Mahsa_Amini | 0 | 1 | 1 | 1 | Individual |
| مکالمه رعنا رحیم پور با مادرش چقدر مسخره و مصنوعی بود سریع هم اومد استوریش کرد که آره منو و جاسوسی بوده ، عزیزمی ها هم بلافاصله عر عر کنان تجزیه میکن خر همون  مادرته جنده  #مهسا_امینی<br>Rana Rahimpour's conversation with her mother was so ridiculous and fake. She quickly posted it on her story, like 'Yeah, it's me, and it's been espionage.' The regime loyalists immediately started braying and analyzing it. The real fool here is your own mother, you whore. #Mahsa_Amini | 0 | 1 | 1 | 1 | Individual - Politics |
| سه ساعته علاف این مامور های گازم اینه از برکات ج/ا که بعد از ۴۳ سال تازه همه شهر  مارو که مرکز استان دارن لوله کشی گاز میکن پول خونم ازمون میگیرن<br>Three hours of gas workers, this is one of the blessings of the GCC. After 43 years, they are installing gas pipes in our city, the provincial capital. They are charging us for our blood money.. | 0 | 1 | 0 | 1 | Idiom |
| از ماست که بر ماست تا به این دین و باورهای بیابان گرد ملخ خوار باور داریم همین آش و همین کاسه<br>It is up to us to believe in this religion and beliefs of the desert, the locust-eating locusts, the same soup and the same bowl. | 0 | 1 | 0 | 1 | Idiom |
| از دی که به سینا به سینا به نجمیه چچه غلطی میکنن از نجمیه هم میخوان به امیر علم حتما بیشرف ها<br>From Day to Sina, from Sina to Najmieh—what the hell are they doing? Now they want to move from Najmieh to Amir Alam. They must be absolute scoundrels. | 0 | 0 | 0 | 0 | Background knowledge |
| هر افغانی چند ریاله؟ بستگی داره چند ساعت میخوای استفاده کنی  قدرت یعنی عبا تن<br>How much is an Afghan worth? It depends on how many hours you want to use them for. | 0 | 0 | 0 | 0 | BG knowledge |
| مسی کردن. نه گوزگوز<br>Power means putting a cloak on Messi, not just farting around.. | 0 | 1 | 0 | 1 | BG knowledge - Idiom |

Table 8: Transfer Learning Model Predictions on Tweets: Samples Illustrating Model Performance Across Different Experiments and Their Strength in Capturing Different Hate Types in Persian.